\documentclass[final,5p,times,twocolumn]{elsarticle}

\usepackage{amssymb}

\usepackage{amsthm}
\usepackage{amsmath}
\usepackage{multirow}

\usepackage{lineno,hyperref}

\usepackage{amssymb}
\usepackage{algorithm}
\usepackage{algorithmic}
\usepackage{epsfig}
\usepackage{subfigure}

\usepackage{float}

\usepackage{longtable}
\usepackage{booktabs}

\usepackage{subfigure}
\usepackage{fancyhdr}
\usepackage{epstopdf}

\usepackage{ulem}
\usepackage{bm}
\usepackage{amsopn}
\usepackage{color,times}
\usepackage{lineno}

\journal{Neurocomputing}

\begin{document}
\begin{frontmatter}



\title{Rethinking ResNets: Improved Stacking Strategies With High Order Schemes}


\author[1,2]{Zhengbo Luo\corref{cor1}%
	}
\ead{lewisluo@fuji.waseda.jp}
\author[1]{Zitang Sun}
\ead{zitang_sun@akane.waseda.jp }
\author[1]{Weilian Zhou}
\ead{zhouweilian1904@akane.waseda.jp}
\author[2]{Zizhang Wu}
\ead{zizhang.wu@zongmutech.com }
\author[1]{Sei-ichiro Kamata}
\ead{kam@waseda.jp}
\cortext[cor1]{Corresponding author}

\address[1]{Graduate School of IPS, Waseda University, Kitakyushu City, Fukuoka, Japan}
\address[2]{Zongmu Technology, Shanghai, China}



\begin{abstract}
Various deep neural network architectures (DNNs) maintain massive vital records in computer vision. While drawing attention worldwide, the design of the overall structure lacks general guidance. Based on the relationship between DNN design and numerical differential equations, we performed a fair comparison of the residual design with higher-order perspectives. We show that the widely used DNN design strategy, constantly stacking a small design (usually 2-3 layers), could be easily improved, supported by solid theoretical knowledge and with no extra parameters needed. We reorganise the residual design in higher-order ways, which is inspired by the observation that many effective networks can be interpreted as different numerical discretisations of differential equations. The design of ResNet follows a relatively simple scheme, which is Euler forward; however, the situation becomes complicated rapidly while stacking. We suppose that stacked ResNet is somehow equalled to a higher-order scheme; then, the current method of forwarding propagation might be relatively weak compared with a typical high-order method such as Runge-Kutta. We propose HO-ResNet to verify the hypothesis of widely used CV benchmarks with sufficient experiments. Stable and noticeable increases in performance are observed, and convergence and robustness are also improved. Our stacking strategy improved ResNet-30 by 2.15 per cent and ResNet-58 by 2.35 per cent on CIFAR-10, with the same settings and parameters. The proposed strategy is fundamental and theoretical and can therefore be applied to any network as a general guideline.
\end{abstract}



\begin{keyword}
	Deep neural networks \sep Neural Ordinary Differential Equations \sep Image processing
	
\end{keyword}

\end{frontmatter}


\section{Introduction}
\label{S1}
Deep neural networks (DNNs) have achieved many exemplary breakthroughs in computer vision, image processing, and signal processing with powerful learning representations from extremely deep structures and massive data. Moreover, repeated simple functions are doing a surprisingly good job while approximating complicated ones; even though this is not fully understood.

Many DNNs have been proposed for more specific tasks, which all perform well. However, despite the tremendous success, we still lack a theoretical understanding of DNNs. In this study, we examine DNNs from a continuous perspective, regard DNNs as discrete dynamical systems, and then improve the stacking strategy of DNNs with high-order numerical methods.

\begin{figure}[!t]
	\centering
	\includegraphics[width=3.4in]{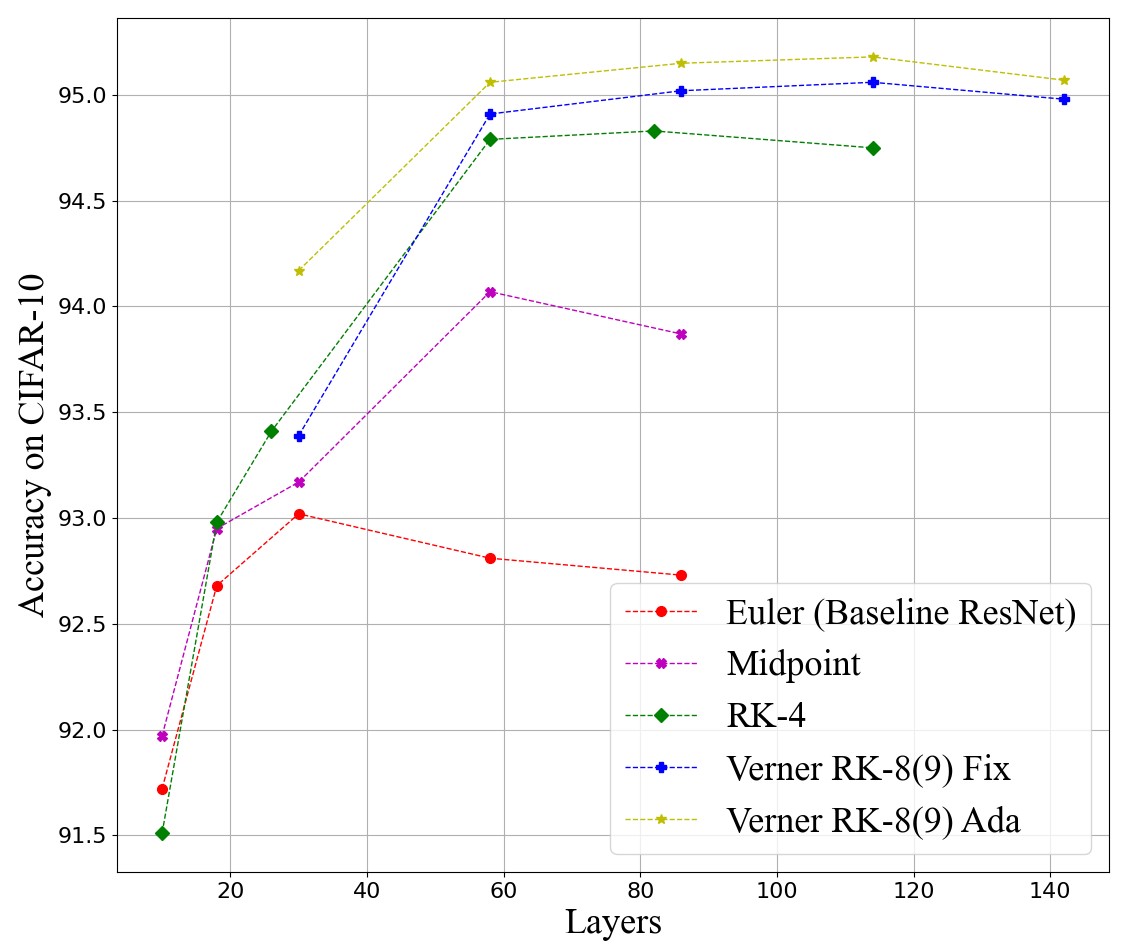}
	\caption{\textit{\textbf{HO-ResNet, stacking the same layers in higher order}}: Given fixed layers, with the same parameters, there are more advanced stacking strategies supported by numerical methods. We implemented three of them to be HO-ResNet, which achieved noticeable improvements in several aspects. Moreover, the degradation problem occurs later with more advanced numerical schemes.}
	\label{Figure-1}
\end{figure}

It is common for DNNs to have more than a hundred layers; however, one usually designs a sub-network with around 2-3 layers and then repeats it multiple times rather than designing a huge one directly. The fixed topology and relationship between blocks lead to the development of dynamic systems.

Enormous possibilities and advantages are gained from this view. For example, mathematical problems are easier to solve and provide direct links to physical sciences and solid theoretical support from differential equations. This study focuses on the depth of DNNs, specifically, the stacking strategy of blocks: constructing DNNs with a given sub-net design in high-order ways.

We show that a certain DNN (ResNet as the example in this study) can be easily improved by slight adjustments in the block stacking strategy, following numerical methods, without any changes in width and depth. Moreover, the improvements are directly proportional to the order of the numerical methods, not only in terms of accuracy but also in terms of convergence and robustness.

\subsection{Related Work}
\label{S2}
The relationship between DNNs and dynamical systems has been widely discussed in recent years. In 2016, Liao et al. \cite{liao} stated the equivalence of ResNet \cite{ResNet} and a specific recurrent network based on formulations from dynamical systems. A systematic proposal was published by Weinan \cite{weinan} in 2017, indicating that DNNs could be thought of as a discretisation of continuous dynamical systems; the proposal and neural ordinary differential equations motivated us to rethink ResNet using advanced numerical methods.

\subsubsection{Neural ODEs}
In 2018, neural ordinary differential equations (NODEs \cite{NODEs}) took a step forward, replacing the discrete DNNs with solvers of ordinary differential equations (ODEs).

To highlight the equivalence, we take a residual network (ResNet) as an example. Let model ResNet be $\mathbf{RES}^{L}(\mathbf{x}): \mathbb{R}^{d_{\text {in }}} \rightarrow \mathbb{R}^{d_{\text {out }}}$ as an $L$-layer residual neural network. We assume that the hidden state $\mathcal{N}$ belongs to $\mathbb{R}^{d_{\text {N}}}$, and the input layer is $\mathcal{N}^{0}(\mathbf{x})=\mathbf{x} \in \mathbb{R}^{d_{\mathrm{in}}}$. The residual design is described as follows:
\begin{equation}
\mathcal{F}\left(\mathbf{x},\left\{\boldsymbol{W}^{i}\right\}\right)+\mathbf{x}, \quad s.t.\quad 1 \leq i \leq L-1.   
\end{equation}
$i$ is an integer only and $\mathcal{F}\left(\mathbf{x},\left\{\boldsymbol{W}^{i}\right\}\right)$ represents the residual mapping to be learned which we do not expand further.

\begin{figure}[!t]
	\centering
	\includegraphics[width=3.4in]{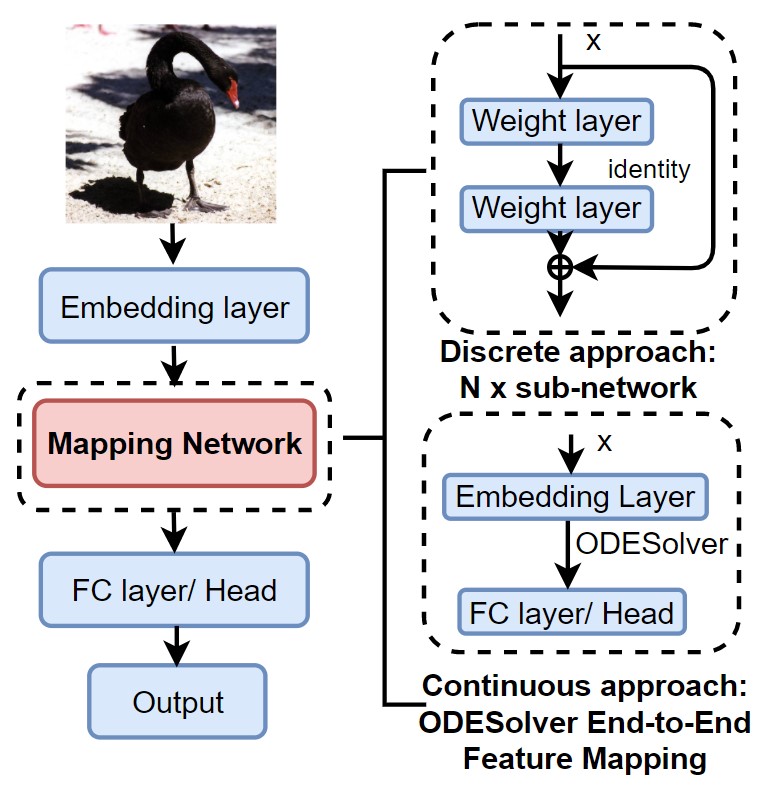}
	\caption{\textit{\textbf{Discrete and continuous approaches of feature mapping}}: Discrete methods usually design a basic structure (2-3 layers), and repeat it N times to have various depths. The size of model, computational cost and performance changes with the depth. Meanwhile, the NODEs map features end-to-end with fully continuous flow and O(1) para size.}
	\label{Figure-2}
\end{figure} 

A sequence of transformation to a hidden state $\mathcal{N}$ from depth $i$ to $i+1$ in residual networks can be written as follows, where $i \in\{1 \ldots L-1\}$ and $\boldsymbol{W}$ refers to the weight matrix:

\begin{equation}
\mathcal{N}^{i+1}=\mathcal{F}\left(\mathcal{N}^{i},\left\{\boldsymbol{W}^{i}\right\}\right) + \mathcal{N}^{i} , \quad s.t.\quad 1 \leq i \leq L-1.  
\end{equation}

The situation above always takes a step $\Delta i = 1$, letting $\Delta i \rightarrow 0$, we have
\begin{equation}
\lim _{\Delta i \rightarrow 0} \frac{\mathcal{N}_{i+\Delta i}-\mathcal{N}_{i}}{\Delta i}=\frac{\mathrm{d} \mathcal{N}(i)}{\mathrm{d} i}=\mathcal{F}(\mathcal{N}(i), i),
\end{equation}
thereby, hidden states can be parameterized using an ODE. Data point $x$ can be mapped into a set of features by solving the initial value problem (IVP):
\begin{equation}
\frac{\mathrm{d} \mathcal{N}(i)}{\mathrm{d} i}=\mathcal{F}(\mathcal{N}(i), i), \quad \mathcal{N}(0)=\mathbf{x},
\end{equation}
at a certain position. $\mathcal{N}(i)$ corresponds to the features learned by the model. NODEs map the input to output by solving an ODE starting from $\mathcal{N}(0)$ and adjust the dynamics to fit the output of the system closer to the label.

\begin{table*}[t]
  \centering
  \begin{tabular}{lll}
    \toprule
    \cmidrule(r){1-3}
    Networks     & Corresponding ODE formula  & ODE Scheme  \\
    \midrule
    ResNet \cite{ResNet} & $\mathcal{N}^{t+1}=\mathcal{N}^{t}+\mathcal{F}\left(\mathcal{N}^{t}, \boldsymbol{W}^{t}\right)$ & Forward Euler\\
    RevNet \cite{revnet} & $\mathcal{Y}^{t}=\mathcal{N}^{t}+\mathcal{F}\left(\mathcal{N}^{t+1}, \boldsymbol{W}^{t}\right)$, $\mathcal{Y}^{t+1}=\mathcal{N}^{t+1}+\mathcal{G}\left(\mathcal{Y}^{t}, \boldsymbol{W}^{t}\right)$ & Forward Euler \\
    PolyNet \cite{zhang2017polynet}  & $\mathcal{N}^{t+1}=\mathcal{N}^{t}+\mathcal{F}\left(\mathcal{N}^{t}, \boldsymbol{W}^{t}\right)+\mathcal{F}\left(\mathcal{F}\left(\mathcal{N}^{t}, \boldsymbol{W}^{t}\right)\right)$ & Backward Euler \\
    LM-ResNet \cite{multistep}  & $\mathcal{N}^{t+1}=\left(1-\Delta t\right)\mathcal{N}^{t}+\Delta t \mathcal{F}\left(\mathcal{N}^{t-1}, \boldsymbol{W}^{t-1}\right)+\mathcal{F}\left(\mathcal{N}^{t}, \boldsymbol{W}^{t}\right)$ & Linear-MultiStep \\
    Second-order CNNs \cite{secondordercnn}  & $\mathcal{N}^{t+1}=2\mathcal{N}^{t}-\mathcal{N}^{t-1}+{\Delta t}^{2}\mathcal{F}\left(\mathcal{N}^{t}, \boldsymbol{W}^{t}\right)$ & Second-order \\
    \bottomrule
  \end{tabular}
  \caption{\textit{\textbf{Connections between neural networks and ODE schemes}}: Some networks have been proposed without awareness of their connection with ODE schemes such as ResNet, RevNet, and PolyNet, while others have been proposed following the guidelines of ODE schemes such as LM-ResNet and Second-Order CNNs. The form of ResNet is widely discussed as the Euler forward scheme, which is the most straightforward way to solve the initial value problem. RevNet is a reversible network, which means that the dynamic can be simulated from the end time back to the initial time. PolyNet includes polynomial compositions that can be interpreted as approximations to one step of the implicit backward Euler scheme. LM ResNet adopted the known linear multistep method in numerical ODEs and achieved the same accuracy by half the parameter size. Moreover, the second-order CNN considers higher-order states, which is another direction that can be further explored.}
  \label{table1}
\end{table*}

In addition to ResNet, several outstanding networks are linked to different numerical schemes (see Table.\ref{table1}). The NODEs are a family of DNNs that can be interpreted as a continuous equivalent of a certain discrete DNN (see Figure.\ref{Figure-2}). Inspired by several results of NODEs, we believe that continuous theoretical concepts could lead to equivalent improvements in discrete DNNs.

\subsubsection{Common high-order numerical schemes for IVP}

With respect to the design of DNNs, structures by NAS have been in a leading position in recent years. However, recent work from the Google team revisited ResNet to conclude that state-of-the-art NAS structures such as Efficient-Nets \cite{effnet} are not necessarily better than ResNets \cite{revisit}. Another excellent piece of work, RepVGG \cite{repvgg}, also achieved state-of-the-art results with a simple VGG-style sub-net. The comeback of ResNet and simple sub-net motivated us to rethink the block-stacking strategy, i.e.,  how to stack these subnets in a better way. We studied ResNet because complicated NAS designs did not seem necessarily better than ResNet.

Euler's scheme is a simple and elegant one-order method, but most networks contain a few basic components. The true expression of the stacked network is nested and in a considerably high order. Will advanced numerical methods make things better? Our stacking designs are based on the corresponding high-order methods for the differential equation. Adopting Euler forward as the baseline ResNet \cite{euler}, mid-point \cite{midpoint}, Runge-Kutta 4th, and Verner's RK-8(9)th Order Embedded Runge-Kutta method \cite{rk4,somerk,rk8} are the general guidelines for our proposed stacking design.

\subsubsection{Effective DNNs that related to ODE schemes}
Table.\ref{table1} shows several outstanding networks with their corresponding schemes. Besides them, many powerful networks are partially related to the ODE scheme. For example, dense connections are like shortcuts of high-order RK schemes \cite{Densenet}, a DenseNet block is nested high-order RK; the One-Shot Aggregation (OSA) proposed by VoVNet \cite{vovnet} is very similar to how RK methods conclude the block output from each state inside. Moreover, in NAS, which discovered several complicated structures, its stacking links (regardless of sub-net differences) may eventually converge to a specific high-order scheme.

\subsection{Contribution}
\paragraph{Motivation} Our strategy is motivated by multiple recent works mentioned above. The comeback of the simple sub-net design made us focus more on the stacking of blocks rather than on the design of a small sub-net. The improved performance of NODEs with advanced solvers indicated to us that it would be of interest to explore stacked discrete networks, following numerical schemes with various orders.

\paragraph{Pros and Cons} Our method enables DNNs to achieve better performance, more stable loss landscape (before and after training), faster convergence, and stronger robustness, which we will show in Section.\ref{S4}. Moreover, one can hardly notice the additional cost when adopting 2-4 order methods, such as mid-point and RK-4, which we will analyse at the end of Section.\ref{S3}. However, the model requires considerable memory to store middle states if fully following a very high-order design, such as Verner's RK-8(9)th scheme.

The main contributions of this study are listed as follows:

\begin{itemize}
	\item We propose the Higher-Order ResNet (HO-Net) to explore, with a given sub-net design, how high-order methods help in terms of performance, convergence and robustness.
	\item We provide sufficient fair comparisons, stacking basic Euler scheme to the same order to enable comparison with high-order methods, therefore no extra parameters and steps.
	\item Visualisations of loss landscapes are given for better understanding.
	\item Corresponding theoretical supports, complexity analysis and sufficient ablation studies.
\end{itemize}

\section{High-Order Residual Networks}
\label{S3}

\begin{figure}[!t]
	\centering
	\includegraphics[width=3.4in]{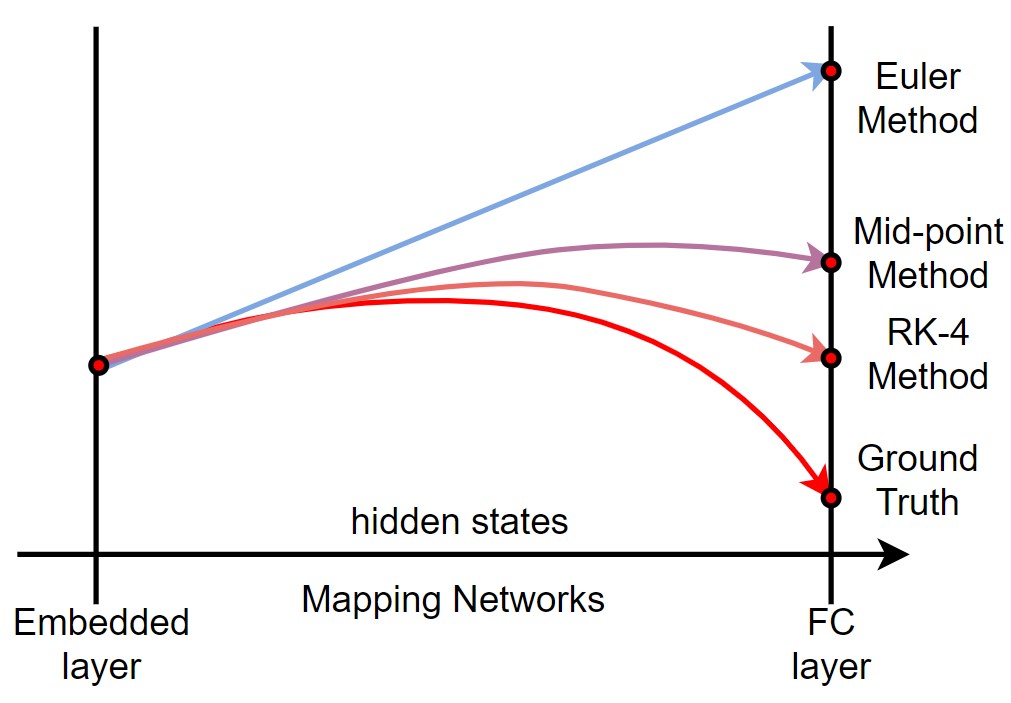}
	\caption{\textit{\textbf{Various schemes with a simple case}}: From end-to-end, higher-order methods with more steps lead to a closer fitting of the target function but also additional steps and costs.}
	\label{Figure-3}
\end{figure}

NODEs compute the gradients of a scalar-valued loss with respect to all inputs of any ODE solver, without back-propagating through the operations of the solver. Similar to different sub-net designs in discrete networks, the methods of the solver make a difference under the same conditions. Thus, advanced solvers make NODEs better, but incur extra computational costs. Figure.\ref{Figure-3} shows a simple case for better understanding.

Will advanced numerical schemes make ResNet better in a discrete scheme with the same cost and parameter size? In this study, baseline ResNet (Euler forward scheme), mid-point scheme, RK-4, fixed Verner-8(9), and adaptive Verner's 8(9) have a basic sub-net with [2, 4, 8, 28, 28] layers, respectively. When comparing ResNet with HO-ResNet, we maintained the same depth and width.

\begin{figure}[!t]
	\centering
	\includegraphics[width=3.4in]{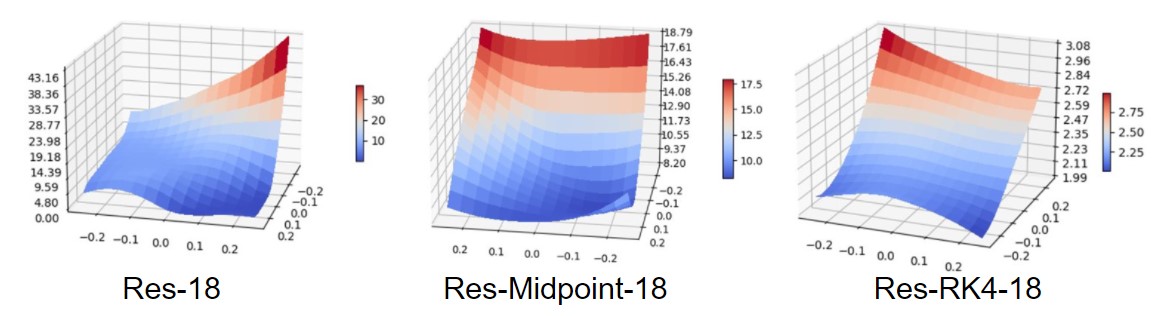}
	\caption{\textit{\textbf{Loss landscape without training}}: When processing the same images, ResNet-18 with Euler, mid-point, and RK-4 schemes have ranges of loss ([0, 43.16], [8.20, 18.79], [1.99, 3.08]), respectively. A high-order scheme significantly improves the stability of models for unseen samples; the loss will not shack too much, which leads to the robustness of the learning rate and other properties, which will be discussed in a later section.}
	\label{Figure-4}
\end{figure}

For example, a ResNet-Euler-18 has eight (8 × 2 layers) Euler-style blocks, one embedding layer, and one FC layer, where ResNet-Midpoint-18 and ResNet-RK4-18 have four (4 × 4 layers) Midpoint-style blocks and two (2 × 8 layers) RK4-style blocks. The higher-order design evidently impacts the network, even without training (see Figure.\ref{Figure-4}, in relation to visualisations of initial conditions.

A few definitions are necessary before introducing HO-ResNet. The situations discussed below are without the embedding layer and FC layers, which means that the input and output are all features. We use $\mathcal{F}$ to describe any sub-net/component, which could be any component such as a VGG-style two-layer sub-net, three-layer bottle net, ViT block, self-attention layer, or simple feed-forward network. 

However, we used a VGG-style block in the experiments because the question 'what form shall $\mathcal{F}$ be?' was not included in our discussion. Our focus is on propagating features in various schemes with a given $\mathcal{F}$. We refer to the input/output features as $\mathcal{N}^{in}$ and $\mathcal{N}^{out}$, denote the middle states of the input as $\mathcal{N}^{mid}$ if there are two layers, $\mathbf{k}_{i}$ is used to indicate the hidden states of the input after a particular $i$th $\mathcal{F}$.

\subsection{Euler Forward Scheme}
Let a network be ResNet-6 which means that two ResBlocks are stacked, and we can write the transforms between the embedding layer and FC layer as follows:
\begin{equation}
\mathcal{N}^{Mid}=\mathcal{F}\left(\mathcal{N}^{in},\left\{\boldsymbol{W}^{i}\right\}\right),
\end{equation}

\begin{equation}
\mathcal{N}^{Out}=\mathcal{F}\left(\mathcal{N}^{Mid} + \mathcal{N}^{in},\left\{\boldsymbol{W}^{i}\right\}\right) + \mathcal{N}^{Mid}.
\end{equation}

$\mathcal{N}^{Mid}$ denotes the output of the first $\mathcal{F}$; one may find that the behaviour of two stacked ResBlocks is very similar to a mid-point ResBlock, as shown in Figure.\ref{Figure-5}.

\subsection{Mid-point Scheme}
The Euler method updates it with the easiest one-step method, which is $\Delta i \mathcal{F}\left(i, \mathcal{N}^{i}\right)$, while the midpoint method updates it in a higher-order manner:

\begin{equation}
\mathcal{N}^{i+\Delta i} = \mathcal{N}^{i} + \Delta i \mathcal{F}\left(i+\frac{\Delta i}{2}, \mathcal{N}^{i}+\frac{\Delta i}{2} \mathcal{N}\left(i, \mathcal{N}^{i}\right)\right).
\end{equation}

\begin{figure}[!t]
	\centering
	\includegraphics[width=3.4in]{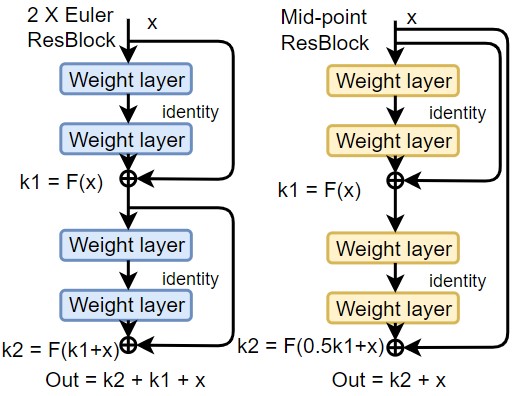}
	\caption{\textit{\textbf{Mid-point ResBlock}}: Comparing two stacked ResBlock-Euler with a single design that following the Mid-point method.}
	\label{Figure-5}
\end{figure}

If we compare ResBlock-Euler and ResBlock-Midpoint, in terms of network design, we will have different stacking, as shown in Figure.\ref{Figure-5}. Rather than a design with two layers, we now have a block design consisting of four layers. Nevertheless, we stack the Euler scheme to the same layers/orders for a fair comparison. 

There were two noteworthy differences: (1). When adding the shortcut to obtain the mid-state, the mid-point design compressed the output of $\mathcal{F}$ by half. (2). the second shortcut is from an earlier location, which is directly from the input. If we write the mid-point ResBlock design as:

\begin{equation}
\mathcal{N}^{Mid}=\mathcal{F}\left(\mathcal{N}^{in},\left\{\boldsymbol{W}^{i}\right\}\right),
\end{equation}

\begin{equation}
\mathcal{N}^{Out}=\mathcal{F}\left(\frac{1}{2}\mathcal{N}^{Mid} + \mathcal{N}^{in},\left\{\boldsymbol{W}^{i}\right\}\right) + \mathcal{N}^{in}.
\end{equation}

One finds that it is very similar to the stacked ResBlock-Euler. Therefore, we assume that stacked ResNet will be easily improved with higher-order methods because numerical problems have solid theoretical backups, which could be more accurate than the lower-order ones.

\subsection{Fourth order Runge-Kutta Scheme}

\begin{figure}[!t]
	\centering
	\includegraphics[width=3.4in]{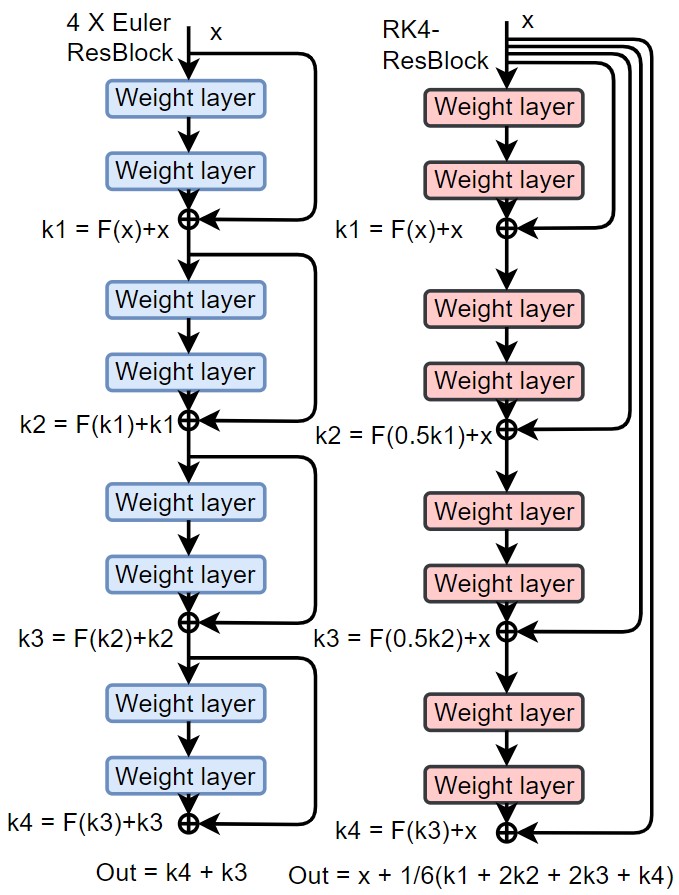}
	\caption{\textit{\textbf{RK4 ResBlock}}: Comparing four stacked ResBlock-Euler with a single design which following the 4th-order Runge-Kutta method.}
	\label{Figure-6}
\end{figure} 

Exploring this further, with a 4th order design, Zhu et al. \cite{rknet} attempted an RK style design. In this study, we use it as a guideline for stacking rather than a certain RK-Net. The Runge-Kutta method (RK4) updates it in a 4-step manner, and our implemented ResBlock-RK4 can be written as follows:

\begin{equation}
\begin{aligned}
\mathbf{k}_{1} &=\mathcal{F}\left(\mathcal{N}^{in}\right), \\
\mathbf{k}_{2} &=\mathcal{F}\left(\frac{1}{2}\mathbf{k}_{1} + \mathcal{N}^{in}\right),\\
\mathbf{k}_{3} &=\mathcal{F}\left(\frac{1}{2}\mathbf{k}_{2} + \mathcal{N}^{in}\right),\\
\mathbf{k}_{4} &=\mathcal{F}\left(\mathbf{k}_{3} + \mathcal{N}^{in}\right),
\end{aligned}
\end{equation}
And the network output is:
\begin{equation}
\mathcal{N}^{Out} =\mathcal{N}^{in}+\frac{1}{6}\left(\mathbf{k}_{1}+2 \mathbf{k}_{2}+2 \mathbf{k}_{3}+\mathbf{k}_{4}\right). 
\end{equation}

Implementing a single ResBlock-RK4 requires eight layers, and thus we compare four stacked ResBlock-Euler with it. The stacked ResBlock-Euler can be written in the same way for easy comparison:

\begin{equation}
\begin{aligned}
\mathbf{k}_{1} &=\mathcal{F}\left(\mathcal{N}^{in}\right),\\
\mathbf{k}_{2} &=\mathcal{F}\left(\mathbf{k}_{1} + \mathcal{N}^{in}\right), \\
\mathbf{k}_{3} &=\mathcal{F}\left(\mathbf{k}_{2} + \mathbf{k}_{1}\right),\\
\mathbf{k}_{4} &=\mathcal{F}\left(\mathbf{k}_{3} + \mathbf{k}_{2}\right),
\end{aligned}
\end{equation}
And the network output is:
\begin{equation}
\mathcal{N}^{Out} =\mathbf{k}_{4} + \mathbf{k}_{3}.
\end{equation}

According to the summary above, we can easily understand the difference between 4th fourth-order RK-ResNet and the four stacked ResBlocks. In Figure.\ref{Figure-6}, we find that RK4-ResBlock is somehow similar to the DenseNet design \cite{Densenet}. However, DenseNet will shortcut every $\mathbf{k}$ to layers after it, which the proposed RK4-ResNet does not. Moreover, HO-ResNet scales the particular output of $\mathbf{k}$ with mathematical support.

Midpoint-ResBlock and RK4-ResBlock only shortcut $\mathbf{k}$ to the block output; identity mappings inside the block are all from the input x of the block. However, some more high-order methods will start to map $\mathbf{k}$ inside the block to other states. However, the coefficient is precisely calculated, rather than simply added.

\subsection{Verner's 8(9)th order Runge-Kutta Scheme}

Of course, one could keep seeking a more advanced scheme to construct ResBlock in a higher-order manner, that contains more layers than just two or three. There are many other versions to guide the network design. Many of these methods are adaptive methods that use a scaling factor. This means that the compression rate of $\mathbf{k}_{i}$ can be adaptive depending on the error returned.

In this study, we implemented the mid-point and Rk4 methods directly, as they are fixed methods. However, we did not fully implement the adaptive Verner's RK-8 \cite{rk8}, as shown in the caption of Figure.\ref{verner} for reasons.

\begin{figure*}
\caption{\textit{\textbf{Verner's RK-8(9)th order scheme}}: The full method has 16 steps and extra processes for error estimation and scale factor adjustment. In our implementation, Verner's 8(9) th-order block has 28 layers instead of (8 × 2 × 2) layers for two reasons: (1). $\mathbf{k}_{15}$ and $\mathbf{k}_{16}$ do not impact the block output $\mathcal{N}^{Out}$ and are used to calculate the adaptive factor $h$. (2). Moreover, one can update $h$ via deep learning frameworks directly by setting it as a learnable tensor value. We compared both fixed $h$ and adaptive $h$, where learnable $h$ is the only extra parameter in HO-ResNet, usually no more than 50, making a rare difference because most models have millions of parameters. $h$ equals one in a fixed Verner block; for the adaptive version, one could use a shard $h$ for all hidden states from $\mathbf{k}_{1}$ to $\mathbf{k}_{14}$ or every state could have a different $h$.}
\centering
\begin{equation}
\begin{aligned}
\mathbf{k}_{1} &=\mathcal{F}\left(\mathcal{N}^{in}\right),\\
\mathbf{k}_{2} &=\mathcal{F}\left(\frac{h}{12}\mathbf{k}_{1} + \mathcal{N}^{in}\right), \\
\mathbf{k}_{3} &=\mathcal{F}\left(\frac{h}{27}\left(\mathbf{k}_{1} + 2\mathbf{k}_{2} \right) + \mathcal{N}^{in}\right),\\
\mathbf{k}_{4} &=\mathcal{F}\left(\frac{h}{24}\left(\mathbf{k}_{1} + 3\mathbf{k}_{3} \right) + \mathcal{N}^{in}\right),\\
\mathbf{k}_{5} &=\mathcal{F}\left(\frac{h}{375}\left(\left(4+94\sqrt{6}\right)\mathbf{k}_{1} - \left(282+252\sqrt{6}\right)\mathbf{k}_{3} + \left(328+206\sqrt{6}\right)\mathbf{k}_{4}\right) + \mathcal{N}^{in}\right),\\
\mathbf{k}_{6} &=\mathcal{F}\left(h\left(\frac{9-\sqrt{6}}{150}\mathbf{k}_{1} + \frac{312+32\sqrt{6}}{1425}\mathbf{k}_{4} + \frac{69+29\sqrt{6}}{570}\mathbf{k}_{5}\right) + \mathcal{N}^{in}\right),\\
\mathbf{k}_{7} &=\mathcal{F}\left(h\left(\frac{927-347\sqrt{6}}{1250}\mathbf{k}_{1} + \frac{7328\sqrt{6}-16248}{9375}\mathbf{k}_{4} + \frac{179\sqrt{6}-489}{3750}\mathbf{k}_{5}+ \frac{14268-5798\sqrt{6}}{9375}\mathbf{k}_{6}\right) + \mathcal{N}^{in}\right),\\
\mathbf{k}_{8} &=\mathcal{F}\left(\frac{h}{54}\left(4\mathbf{k}_{1} + \left(16-\sqrt{6}\right)\mathbf{k}_{6} + \left(16+\sqrt{6}\right)\mathbf{k}_{7}\right) + \mathcal{N}^{in}\right),\\
\mathbf{k}_{9} &=\mathcal{F}\left(\frac{h}{512}\left(38\mathbf{k}_{1} + \left(118-23\sqrt{6}\right)\mathbf{k}_{6} + \left(118+23\sqrt{6}\right)\mathbf{k}_{7} - 18\mathbf{k}_{8}\right) + \mathcal{N}^{in}\right),\\
\mathbf{k}_{10} &=\mathcal{F}\left(h\left(\frac{11}{144}\mathbf{k}_{1} + \frac{266-\sqrt{6}}{864}\mathbf{k}_{6} + \frac{266+\sqrt{6}}{864}\mathbf{k}_{7}- \frac{1}{16}\mathbf{k}_{8 }- \frac{8}{27}\mathbf{k}_{9}\right) + \mathcal{N}^{in}\right),\\
\mathbf{k}_{11} &= \mathcal{F}\left(h\left(\frac{5034-271\sqrt{6}}{61440}\mathbf{k}_{1} + \frac{7859-1626\sqrt{6}}{10240}\mathbf{k}_{7} + \frac{813\sqrt{6}-2232}{20480}\mathbf{k}_{8}+ \frac{271\sqrt{6}-594}{960}\mathbf{k}_{9} + \frac{657-813\sqrt{6}}{5120}\mathbf{k}_{10}\right) + \mathcal{N}^{in}\right),\\
\mathbf{k}_{12} &= \mathcal{F}\left(h\left(-8.14164\mathbf{k}_{1} -574.436\mathbf{k}_{6} + 847.88\mathbf{k}_{7}+ 113.719\mathbf{k}_{8} + 626.94\mathbf{k}_{9} +
605.73\mathbf{k}_{10} - 328.69 \mathbf{k}_{11}\right) + \mathcal{N}^{in}\right),\\
\mathbf{k}_{13} &= \mathcal{F}\left(h\left(0.0878\mathbf{k}_{1} + 0.69337\mathbf{k}_{6} - 1.9\mathbf{k}_{7}+ 0.23\mathbf{k}_{8} - 0.69\mathbf{k}_{9} -
0.077\mathbf{k}_{10} + 2.49 \mathbf{k}_{11} + 0.0018 \mathbf{k}_{12}\right) + \mathcal{N}^{in}\right),\\
\mathbf{k}_{14} &= \mathcal{F}\left(h\left(-0.1\mathbf{k}_{1} + 5.575\mathbf{k}_{6} + 7.486\mathbf{k}_{7} - 6.23\mathbf{k}_{8} + 2.27\mathbf{k}_{9} -
4.89\mathbf{k}_{10} - 4.86 \mathbf{k}_{11} - 0.0235 \mathbf{k}_{12} + 1.78\mathbf{k}_{13} \right) + \mathcal{N}^{in}\right),\\
\end{aligned}
\end{equation}
And the network output is:
\begin{equation}
\mathcal{N}^{Out} = \mathcal{N}^{in} + h \left(0.06\mathbf{k}_{1} - 0.19\mathbf{k}_{8} + 0.72\mathbf{k}_{9} - 0.72\mathbf{k}_{10} +0.75\mathbf{k}_{11} +0.0004\mathbf{k}_{12} + 0.34\mathbf{k}_{13} + 0.032\mathbf{k}_{14} \right)
\end{equation}
\label{verner}
\end{figure*}

The full Verner's RK-8(9) ResBlock design shall contain 16 $\mathcal{F}$ which means 32 layers. We use $\mathbf{k}_{i}$ to represent the output of each block, where $i$ is an integer in the range [1-16]. Unfortunately, no flowchart is provided because of the complicated connections between the hidden states.

The final output of the Verner-8(9) ResBlock depends only on the shortcut from the start and $\mathbf{k}_{i}$, where $i$ in [1,8,9,10,11,12,13,14]. The scale factor $h$ should be adjusted depending on $\mathbf{k}_{i}$, where $i$ in [1,8,9,10,11,12,13,14,15,16]. Thereby, one can implement a fixed version with an out-of-scale factor adjustment using 14 $\mathcal{F}$. However, $\mathbf{k}_{15}$ and $\mathbf{k}_{16}$ are required to update $h$ if one wants to fully follow Verner's method, in which we update $h$ by directly setting it to be learnable, in order to compare fixed and adaptive versions with the same computational cost.



.



\subsection{Complexity Analysis}
We compare the situation in which the same size, number, and design of the layers are used. The space complexity can be described by the \begin{equation}
\text { Space } \sim O\left(\sum_{i=1}^{D} K_{i}^{2} \cdot C_{l-1} \cdot C_{i}+\sum_{i=1}^{D} M^{2} \cdot C_{i}\right)
\end{equation}
where K is the kernel size, C refers to the number of channels, and D denotes the depth of the network. Thus, the complexity of various ResNet designs is the same, with no extra parameters, theoretically. However, in practice, extra space is required to maintain the output inside the high-order block because the block output depends on more states at particular layers rather than two fixed positions, compared with the baseline ResNet.

Specifically, baseline ResNet only maintains a shortcut from the block input, as does the mid-point block. For the Runge-Kutta 4 block, we need three more shortcuts for the block output. However, this could be optimised to one more shortcut.

For Verner's RK-8 block, this situation becomes rather complicated. From the state $\mathbf{k}_{5}$, every state starts to depend on other previous states, besides the input from the last state; we shall maintain 16 more shortcuts for the full version or 14 more shortcuts for our version. Furthermore, the space requirement is extremely high while forwarding, which is a crucial shortage; sufficient space should be provided when using such a design. Therefore, it is not recommended to use Verner's block; applying the adaptive step from Verner's block to a fourth-order scheme would be a better solution.

Our design does not affect the first term, $\sum_{i=1}^{D} K_{i}^{2} \cdot C_{l-1} \cdot C_{i}$, but only the second term, $\sum_{i=1}^{D} M^{2} \cdot C_{i}$.
Baseline ResNet maintains only one feature map for shortcut as $\sum_{i=1}^{D} M^{2} \cdot \left(C_{i} + 1\right)$, similar to the midpoint block. RK4-Block and Fixed-RK8-Block will take $\sum_{i=1}^{D} M^{2} \cdot \left(C_{i} + 2\right)$ and $\sum_{i=1}^{D} M^{2} \cdot \left(C_{i} + 15\right)$, respectively. We could say that the extra space requirement could be ignored generally for second-fourth-order schemes but is slightly too large for full Verner's RK-8(9).

We may describe the time complexity by:
\begin{equation}
\text { Time } \sim O\left(\sum_{i=1}^{D} M_{i}^{2} \cdot K_{i}^{2} \cdot C_{i-1} \cdot C_{i}\right)
\end{equation}

According to the explanation above, we know that HO-ResNet will frequently scale the shortcuts, unavoidably creating extra multi/add processes. Compared with the baseline ResNet, the midpoint block has one additional multiplication operation for every four layers. Therefore, we shall add $\frac{1}{4}\sum_{i=1}^{D} M^{2} \cdot C_{i}$.

The RK-4 block requires six more multiplication operations and three additional operations every eight layers. Therefore, we shall add $\frac{9}{8}\sum_{i=1}^{D} M^{2} \cdot C_{i}$. It is difficult to observe the complexity changes among Euler, Midpoint, and RK4. However, Verner's RK-8 requires a considerable number of storage spaces because of the highly correlated hidden states and extensive additional multi/add scaling. The time consumption is approximately 1.15x-1.2x while the previous three remain the same.

\begin{table}[H]
    \centering
    \begin{tabular}{ccccc}
    \toprule
        Schemes & Euler & Midpoint & RK-4 & Verner-8(9)\\
        \midrule
        Memory & 1x & 1x & 1.5x(1x) & 7x \\
        \midrule
        \midrule
        Time & 1x & 1x & 1x & 1.15x-1.2x \\
        \bottomrule
    \end{tabular}
    \caption{\textit{\textbf{Multiple relationships of memory requirements and running time:}} The second to fourth-order scheme has rare extra operations, and thus does not reflect on the running time. However, more memory is required because RK-4 blends all hidden states for the output. Verner's 8(9) requires huge amounts of memory because it blends not only hidden states for output but also previous states for hidden states in a later stage.}
    \label{table2}
\end{table}

Table.\ref{table2} shows the multiple relationships of memory and running time in practice, taking the ResNet as the baseline. There is no room to optimize the huge cost of Verner's 8(9), but one could remove the blending of RK-4 and output the $\mathcal{N}^{in} + \mathbf{k}_{4}$ only instead of $\mathcal{N}^{in}+\frac{1}{6}\left(\mathbf{k}_{1}+2 \mathbf{k}_{2}+2 \mathbf{k}_{3}+\mathbf{k}_{4}\right)$. RK-4 will have the same memory requirement as baseline Euler by doing so with no evident performance drop.

\begin{figure}[!t]
	\centering
	\includegraphics[width=3.45in]{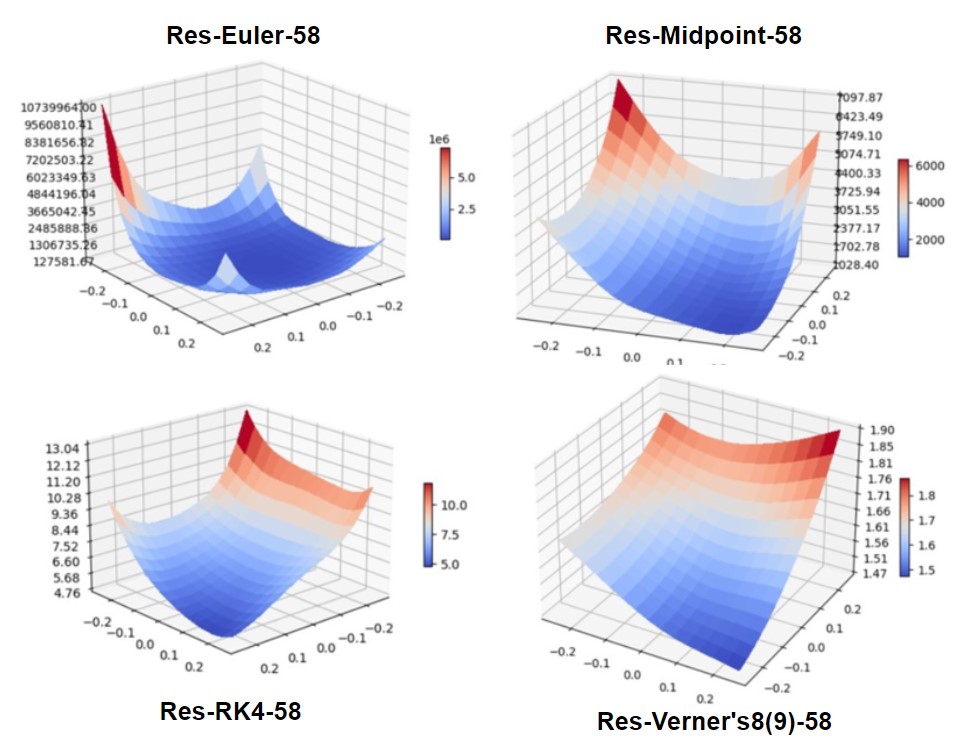}
	\caption{{\bfseries\textit{Visualisation of untrained ResNet-58 with various schemes :}} The range of loss exploded with a low-order scheme. Euler, Midpoint, RK-4, and Verner's 8(9) have loss ranges ([1e5, 1e7], [1028, 7097], [4.76, 13.04], [1.47, 1.90]). The stability is significantly improved with higher order, which means that gradient problems, such as vanishing or exploding, are less likely to occur.}
	\label{Figure-8}
\end{figure}

\begin{figure}[!t]
	\centering
	\includegraphics[width=2.4in]{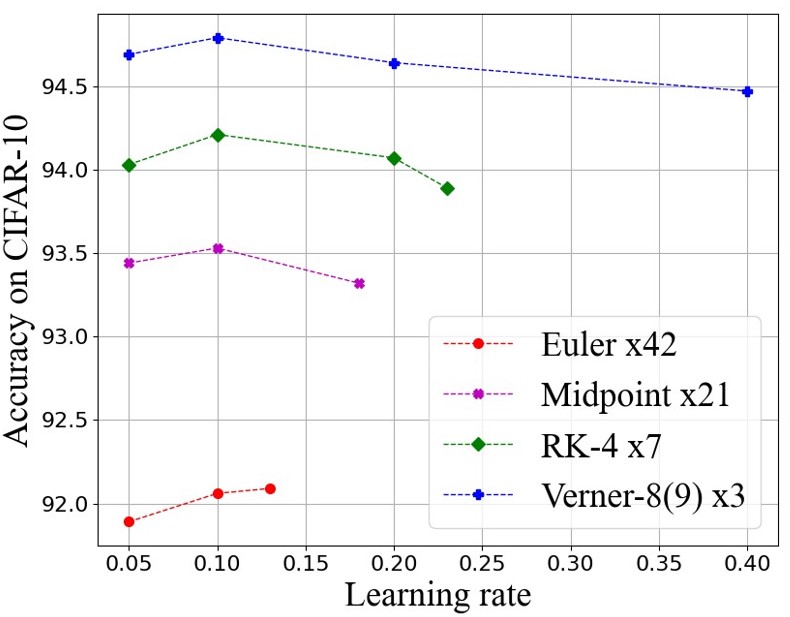}
	\caption{{\bfseries\textit{ResNet-86 with various schemes against different initial learning rates :}} Under the same condition, ResNet with higher-order could be trained with a noticeable larger learning rate. Baseline Euler scheme diverged with learning rate 0.13 and all lr larger than it, while Mid-point could converge with lr close to 0.2, RK-4 could converge at lr 0.23 and Verner's-8(9) can converge at lr 0.4 (or even above).}
	\label{Figure-9}
\end{figure}

\begin{figure}[!t]
	\centering
	\includegraphics[width=3.45in]{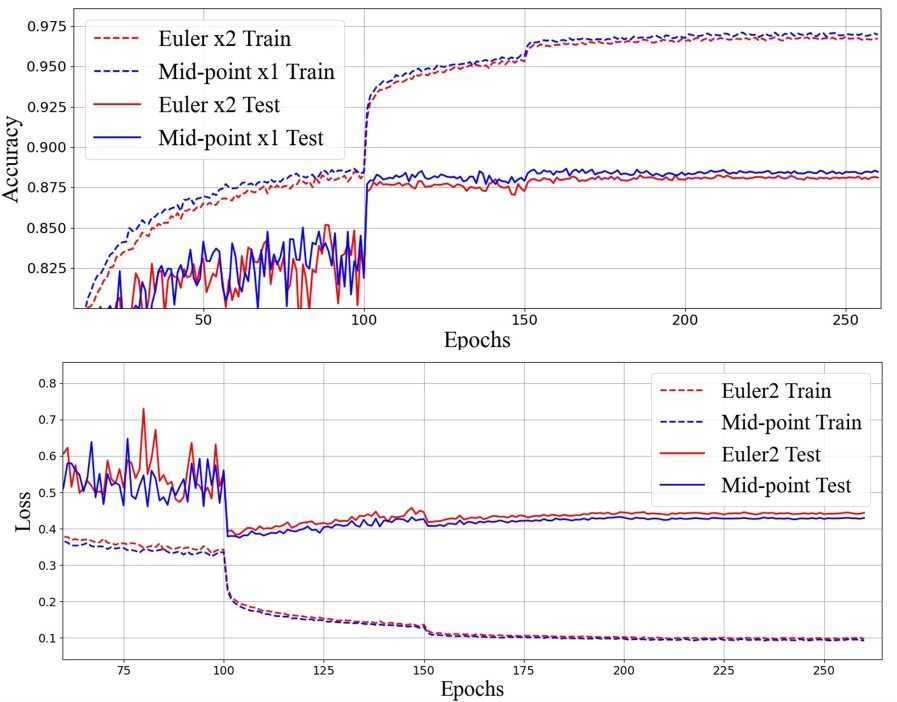}
	\caption{{\bfseries\textit{One midpoint block vs. two Euler blocks:}} The second-order scheme shows noticeable improvements in both accuracy and loss during training and testing.}
	\label{Figure-10}
\end{figure}

\begin{figure}[!t]
	\centering
	\includegraphics[width=3.45in]{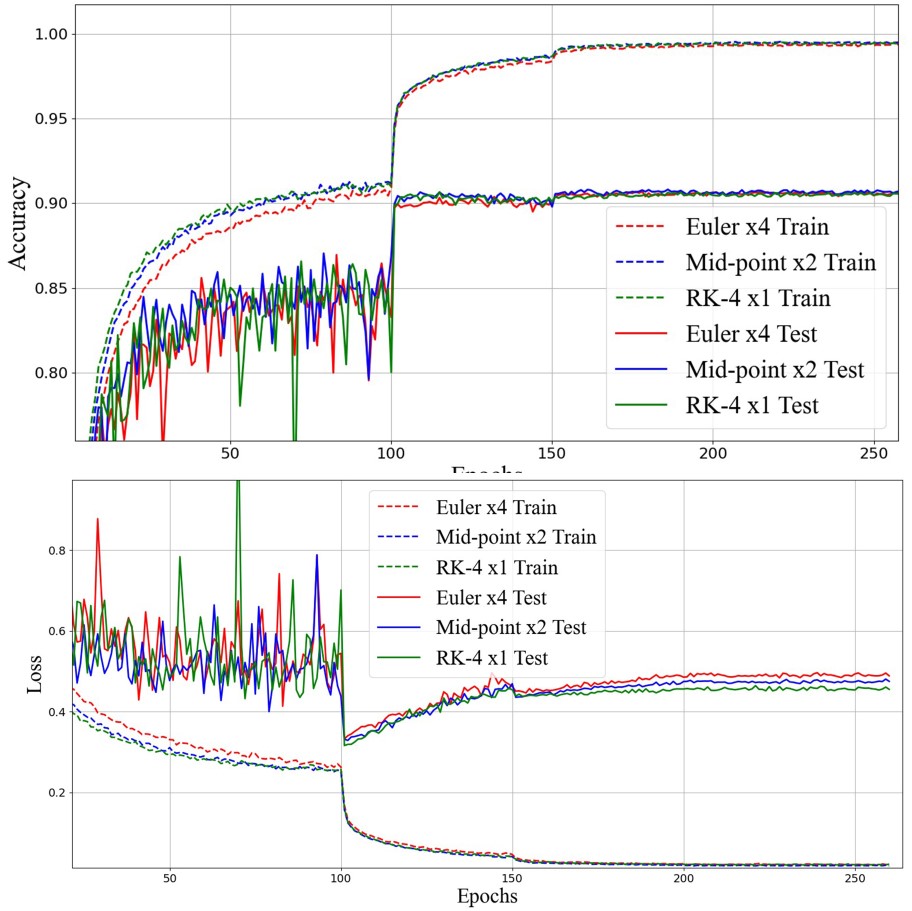}
	\caption{{\bfseries\textit{One Rk4 block vs. two Midpoint blocks and four Euler Blocks:}} The fourth-order RK-4 scheme outperforms the mid-point and baseline Euler as a basic sub-net. Note that the advantages will become increasingly evident for deeper situations.}
	\label{Figure-11}
\end{figure}

\begin{figure*}[!t]
	\centering
	\includegraphics[width=7in]{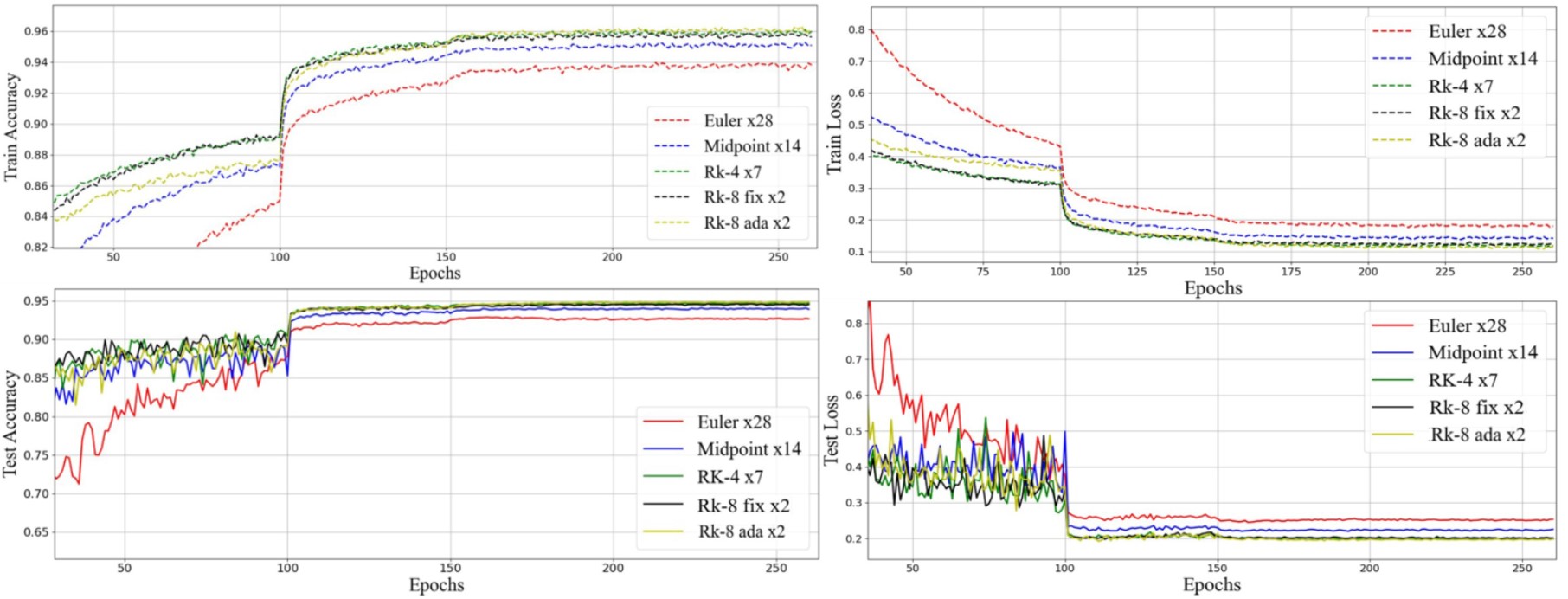}
	\caption{{\bfseries\textit{ResNet-58 with various schemes:}} Training accuracy and loss on top and testing accuracy and loss at the bottom. Note that the difference between schemes is becoming increasingly noticeable compared with observations on shallow versions.}
	\label{Figure-12}
\end{figure*}

\begin{figure}[!t]
	\centering
	\includegraphics[width=2.8in]{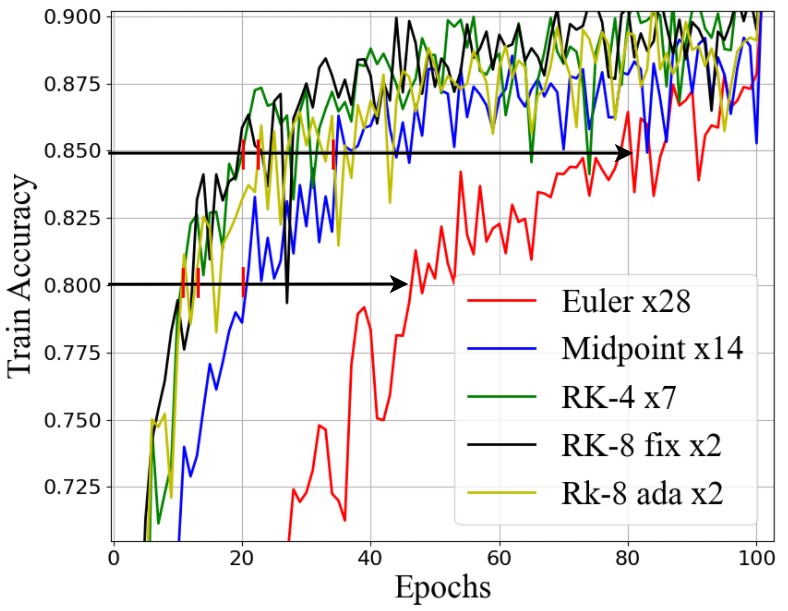}
	\caption{{\bfseries\textit{Running Time comparison approaching 0.8, and 0.85 train accuracy (ResNet-58):}} High-order schemes have evident advantages in terms of the speed of convergence. The second-order mid-point takes half of the time to reach 0.8, and 0.85, compared with Euler, while the fourth and 8(9) order methods take 20–30 per cent of the time. See details in Table.\ref{table4}. }
	\label{Figure-13}
\end{figure}


\section{Experiments}
\label{S4}
\subsection{Datasets and implementation}
In this study, we conducted most studies on the CIFAR–10 datasets \cite{cifar10}, which all consist of coloured natural images with 32 × 32 pixels with 50,000 images for training and 10,000 images for testing, which has 10 target classes in total for CIFAR-10 with 6,000 images per class.

We directly stacked blocks 64 dim after a conv(3, 64). Stochastic gradient descent \cite{SGD} under a weight decay of 1e-4 and momentum of 0.9, was used with weight initialisation in \cite{13} and batch norm \cite{15}. The models were trained with a batch size of 128, starting with an initial learning rate of 0.1, step scheduler with [100, 150, 200, 230], factor 0.1, and 260 epochs trained. 

\subsection{Augmentations and tricks}
Our focus was on the differences in behaviour for each of the stacking strategies to verify whether the high-order stacking could provide general guidance for neural network design, but not on pushing a certain line of results or seeking a more considerable number. Therefore, we intentionally compared those stacking strategies using methods that enabled fair comparisons to be drawn.

When comparing a single block design with the corresponding deep baseline ResNet, we used raw data and observed overfitting even under 10 layers. Therefore, we used the auto augments' CIFAR10 policy \cite{autoaug} to address this question.

\subsection{Robustness against learning rate}
Figure.\ref{Figure-8} shows a deeper situation with different schemes. Comparing with the case of ResNet-18 shown by Figure.\ref{Figure-4}, the loss exploded due to the chain rule. Meanwhile, higher order schemes show significant advantages, such as more flatten loss surfaces being provided. Such a phenomenon greatly enhances the robustness of models when different learning rates are given.

The learning rate is one of the primary hyperparameters of deep-learning systems. Slight adjustments in the learning rate could cause vast impacts on the whole system, large learning rates will easily lead to divergence, and small learning rates usually fail to converge well.

Taking ResNet-86 as an example, 84 layers could consist of [42, 21, 7, 3] of [Euler, Midpoint, RK-4, Verner's 8(9)] blocks. Figure.\ref{Figure-9} shows how higher-order schemes are dealing with a larger range of learning rates. ResNet-86 in the 4-th/8-th order scheme could be well trained with 2x/4x learning rate at the point where the baseline Euler diverged

\begin{table}[H]
    \centering
    \begin{tabular}{cccc}
    \toprule
        Models & Layers & Params & cifar-10 \\
        \midrule
         Euler X4 & 10 & 0.3M & 91.72 \\
         Midpoint X2 & 10 & 0.3M & \bfseries91.97  \\
         RK4 X1 & 10 & 0.3M & 91.51 \\
         \midrule
         Euler X8 & 18 & 0.59M & 92.68 \\
         Midpoint X4 & 18 & 0.59M & 92.95 \\
         RK4 X2 & 18 & 0.59M & \bfseries92.98 \\
         \midrule
         Euler X14 & 30 & 1.03M & 93.02 \\
         Midpoint X7 & 30 & 1.03M & 93.17 \\
         RK8 X1 Fix & 30 & 1.03M & 93.39 \\
         RK4 X3 & 26 & 0.89M & 93.41 \\
         RK8 X1 Adaptive & 30 & 1.03M & \bfseries94.17 \\
         \midrule
         Euler X28 & 58 & 2.07M & 92.81 \\
         Midpoint X14 & 58 & 2.07M & 94.08\\
         RK4 X7 & 58 & 2.07M & 94.79 \\
         RK8 X2 Fix & 58 & 2.07M & 94.91\\
         RK8 X2 Adaptive & 58 & 2.07M+2 & \bfseries95.06\\
         \midrule
         Euler X42 & 86 & 3.01M & 92.73 \\
         Midpoint X21 & 86 & 3.01M & 93.87\\
         RK4 X10 & 82 & 2.87M & 94.83 \\
         RK8 X3 Fix & 86 & 3.01M & 95.02\\
         RK8 X3 Adaptive & 86 & 3.01M+3 & \bfseries95.15\\
         \bottomrule
    \end{tabular}
    \caption{\textit{\textbf{Classification accuracy on CIFAR-10}}: Various schemes with different depth are compared. All schemes perform similar when depth smaller than 18. After that, higher order schemes are showing more and more apparent advantages in terms of classification accuracy. }
    \label{table3}
\end{table}

\subsection{Fair comparison of single block}
If one compares each block design with equivalent deep baseline ResBlocks, it is found that the improvements are stable but not apparent in the shallow situation. However, the advantages were rapidly becoming evident in more profound cases.

\paragraph{Euler-x2 and Midpoint-x1}
Given ResNet-6, we have four layers beside the input and output layers. We performed experiments on CIFAR-10 raw data with rare tricks; as shown in Figure.\ref{Figure-10}, loss and accuracy are stably improved.

\paragraph{Euler-x4, Midpoint-x2 and RK4-x1}
If ResNet-10 was given, one could stack four baseline ResBlocks, which is the red line in Figure.\ref{Figure-11}, while Midpoint-2 and single RK-4 block benefit from more accurate updating methods.

\subsection{Deeper Situation with augmentation}
We noticed overfitting when using ResNet shallower than 10. Therefore, we added auto-augmentation \cite{autoaug} in this comparison. Without the input and output layers, 56 layers could be divided into Baseline ResBlock X28, Midpoint x14, RK4-Block x7, and fixed-RK8 X2. 

As shown in Figure.\ref{Figure-12}, baseline ResNet, represented by the red line, is clearly improved by high-order stacking methods, using the same setting and parameters. We noticed that the fixed-RK8 method has the potential to improve. In the early stages it was above RK-4. In Verner's RK-8(9) \cite{rk8} design, one should also maintain a scale factor $h$ and $error$ to adjust the step side. Adaptive-step Verner's RK-8(9) converges at a slightly slower rate but  outperforms all other schemes overall.

Table.\ref{table3} shows more details against different depths; all schemes perform similarly in shallow cases but show more evident gaps while getting deeper. Moreover, there are numerous benefits to be gained from high-order, besides the performance, such as loss landscape and robustness against hyperparameters, Figure.\ref{Figure-13} and Table.\ref{table4}  show the improvement in the speed of converging.

\begin{table}[H]
    \centering
    \scalebox{0.9}{
    \begin{tabular}{cccccc}
    \toprule
        Schemes & Euler & Midpoint & RK-4 & Fix RK-8 & Ada RK-8\\
        \midrule
        To 0.8 & 1x & 2.27x & 4.43x & 3.31x & 3.84x \\
        To 0.85 & 1x & 2.28x & 3.92x & 3.53x & 3.03x \\
        \midrule
        Accuracy & 92.71 & 94.08 & 94.79 & 94.72 & 95.06 \\
        \midrule
        Extra para & / & / & / & / & 2 \\
        \bottomrule
    \end{tabular}}
    \caption{\textit{\textbf{Ratio of time cost (ResNet-58):}} Note that Verner's RK-8 costs a 1.15 increase in time compared to the others; this situation leads RK-4 to be deemed to achieve the best performance in terms of time. Adaptive Verner's RK-8 appears to struggle in the mid-stage but does converge to achieve the best overall performance in the end, with only two extra parameters needed.}
    \label{table4}
\end{table}

\subsection{Scaling Factors}
The most effective DNNs have a relatively light stacking design between blocks and rarely feature scaling factors on block outputs. Following the Midpoint and RK4 scheme, the output should be compressed with a scaling factor of 0.5, and RK4 will have 1/6 and 1/3 factors when averaging the output inside the block.

In RK-8 \cite{rk8}, a constraint [0.125, 4] was given while scaling the output from a particular layer. Although we did not further explore the adaptive stacking design, we tried a simple test on how various scaling factors impact the same DNNs.

Taking ResNet-58 as an example, which has two fixed Verner's RK-8 blocks and two extra parameters, see Table.\ref{table4}. Although the full Verner's RK-8 constrains the factor inside [0.125, 4], we did not follow the constraint when applying the adaptive step for practical reasons. However, the results still show a noticeable improvement, as expected.

\subsection{Discussing}
\paragraph{Stacking strategy is vital}
In the past few years, some network designs have been regarded as ODE schemes such as ResNet and PolyNet \cite{ResNet,Poly}; some networks are designed under the guidance of ODE schemes such as LM-ResNet and RK-Net \cite{multistep,rknet}. However, most of them focus more on the subnet design, while the stacking strategy is also critical. Block-to-block is similar to layer-to-layer, especially when some networks will even stack 84 blocks in a single stage.

\paragraph{Three principles from High order schemes}
(1). One shall apply an adaptive time step; it can easily be plugged in to any network. In fact, several excellent studies have already proposed methods in this manner from other perspectives. Highway networks \cite{highway}, Fixup \cite{fixup}, SkipInit \cite{skipinit}, ReZero \cite{re0}, and LayerScale \cite{going} are the approaches that are most closely related. (2). One should consider connections over long distances, but not nested ones such as DenseNet \cite{Densenet} to avoid huge memory requirements. (3). Blending from different hidden states works, but it is not as important as the previous two points. Moreover, it requires at least 1x extra memory.

\paragraph{Unavoidable degradation} ResNet \cite{ResNet} significantly alleviated the degradation problem that deeper networks shall perform better in theoretical but worse in practice. The shortcut does not provide extra capacity but makes models easier for training, so does HO-ResNet. We implemented ResNet with advanced numerical methods based on the equivalence between ResNet and Euler Forward scheme; the results in Figure.\ref{Figure-1} show that the degradation was better handled and further alleviated with more advanced numerical schemes; however, the degradation will still come eventually.

\section{Conclusion}
\label{S5}
Inspired by several existing studies \cite{weinan,multistep,NODEs,revisit,ResNet}, we assume that DNNs are extremely complicated numerical solvers for differential equations. Most designs are relatively low-order, while the networks are frequently deeper than 100 or 200. We propose a general higher-order stacking strategy, following several numerical methods, which proved that high-order numerical methods could be adopted as general guidance for block stacking and network design. Sufficient experiments show that high-order improvements are stable, robust, and fully explainable in terms of math. We show that with very few changes in stacking, ResNet has shown remarkable durability.






\bibliographystyle{elsarticle-num} 
\bibliography{ref}

\end{document}